# HENRI: High Efficiency Negotiation-based Robust Interface for Multi-party Multi-issue Negotiation over the Internet

Saurabh Deochake
Department of Information Technology
Maharashtra Institute of Technology
Pune 411038, India
(+91)-8983466534
saurabh.deochake
@gmail.com

Shashank Kanth
Department of Information Technology
Maharashtra Institute of Technology
Pune 411038, India
(+91)-9221958582
kanthshashank
@gmail.com

Subhadip Chakraborty
Department of Information Technology
Maharashtra Institute of Technology
Pune 411038, India
(+91)-8097693404
subhadip.schakraborty
@gmail.com

Suresh Sarode
Department of Information Technology
Maharashtra Institute of Technology
Pune 411038, India
(+91)-9765119772
sarodesuresh.r
@gmail.com

Vidyasagar Potdar
School of Information Systems
Curtin University
Perth 6845, Australia
(+61)-421447048
v.potdar@curtin.edu.au

Debajyoti Mukhopadhyay
Department of Information Technology
Maharashtra Institute of Technology
Pune 411038, India
(+91)-7709152655
debajyoti.mukhopadhyay
@gmail.com

## ABSTRACT
This paper proposes a framework for a full fledged negotiation system that allows multi party multi issue negotiation. It focuses on the negotiation protocol to be observed and provides a platform for concurrent and independent negotiation on individual issues using the concept of multi threading. It depicts the architecture of an agent detailing its components. The paper sets forth a hierarchical pattern for the multiple issues concerning every party. The system also provides enhancements such as the time-to-live counters for every advertisement, refinement of utility considering non-functional attributes, prioritization of issues, by assigning weights to issues.

## Categories and Subject Descriptors
H.3.3 [**Information Search and Retrieval**]: Information filtering, Retrieval models.

## General Terms
Algorithms, Performance, Design, Experimentation, Human Factors, Theory, Verification

## Keywords
Multi-party multi-issue, Utility Function, Multi threading, Master coordinator and Coordinator object, Concurrent negotiation

## 1. INTRODUCTION
Negotiation is a process in which different entities concur to an agreement on a joint future initiative or behavior. The substantial rate of increase in the number of transactions executed through electronic channels such as the internet in the B2B and the B2C E-commerce has led to the need of better negotiation models and protocols. The current E-business activities are no more restricted to the enterprise but it considers various schemes such as advertisements of products outlining their features and their categories in the internet, providing scopes for carrying out offers and counter offers for each of the parties [11-13]. We provide a model that can be used in a wide variety of negotiation scenarios including B2B and B2C e-commerce activities.

This model is designed to be used by automated agents (or more specifically, automated buyer and seller agents in our case). An agent is an encapsulated computer system based on an unspecified environment with the capability of flexible, autonomous action in that environment in order to meet its design activities.

Negotiation objects ranges from single issues, viz., cost to multiple (quality, profit), etc. For a deal to be finalized, all the issues which have been arranged hierarchically in our model have to be satisfied.

Our model provides us with a concurrent solution to the execution of different negotiations at the same time i.e., multiple agents negotiate over multiple issues at the same time in parallel. A linear mathematical paradigm for proposing counter offers is also suggested. The negotiation protocol follows priorities of the issues, taking into perspective a utility function which considers the non functional attributes and assigns weight in accordance to the relative importance of the features.

The paper is structured as follows: Section 2 gives an outline of the existing work in this field, Section 3 describes of the model proposed in this paper, Section 4 includes the salient features of our model, Section 5 has a simulation for our model, Section 6 suggests future work and finally, we conclude this paper in section 7.

## 2. RELATED WORK

An overview of the process of negotiation can be represented by a buyer and a seller agent with their respective deadline tmax by which the agreement has to be reached, for finalizing the deal. Multiple issues such as price, quality, etc. have to be considered and once the buyer or the seller finalizes the deal, it declines offers from all others. Three approaches have been used: Game theory, Heuristics and Argumentation Approach.

Game theory [2] attempts to capture behavior in strategic behavior situations, where each player's success in making a decision depends on the decision of others. A protocol based on this approach is simple to implement but the biggest problem with this is that it is computationally complex. The number of calculations increases exponentially in each round. Another problem is that it is only suitable for specialized models that are used for specific types of interdependent decision making.

Some of the limitations of game theory based techniques can be overcome by making use of heuristics [1]. Here, the agreement space can be searched in a non-exhaustive fashion. Heuristics are experience based techniques that make educated guesses. So, what we get are good results and not optimal results. The problems with using heuristics are that, (1) the outcomes are sub-optimal as they do not consider the full space of possible outcomes and (2) it is impossible to predict how the system and its constituent agents will behave in different circumstances so it requires extensive evaluation.

Finally, Argumentation Framework [7] applies logic based reasoning and argumentation in the proper design of agents for web services.

Thus, we propose a negotiation model that attempts to arrive at a compromise among the above three approaches and is computationally simple at the same time.

To implement an efficient negotiation system, there should be capable subcomponents of the negotiation system. The main types of the negotiation services that could be included are as given in [6]:

(1) Fully-delegated negotiation services: "Fully delegated" can be defined as a term that prescribes the incorporation of automated negotiation agents in the system. (2) Interoperable messaging system: This messaging system has the possibility of being the message broker. (3) Negotiation process management: The negotiation process management forces the negotiation rules that should be carried out while in the negotiation.

In order to aid an open and dynamic negotiation environment [9], the requirements for a product specification language are: (1) Formal Semantics: There is a requirement of a language with formal semantics to accommodate heterogeneity of participants. (2) Dynamic re-configurability: Dynamic changes are allowed to be incorporated into the syntax through this feature. (3) Compatibility: Protocols can be designed to follow a modular style using this. Compatibility permits the construction of negotiation protocols that incorporate independent negotiation. (4) Composability using runtime leads to the option of dynamically adding new protocols switching among them. Our model allows fully-delegated negotiation services by using automated agents, negotiation process management and formal semantics by enforcing fixed communication primitives and compatibility by defining a uniform interface which can be implemented using web services.

Interaction among agents requires a uniform messaging system or a standard interface which will allow communication using fixed message primitives. T. D. Nguyen and N. R. Jennings in their paper for concurrent, bilateral negotiation [3] make use of the following message primitives: Offer (a proposal made by one agent to the other), *Counter-offer* (a revised proposal from an agent in response to a proposal it has received), *Accept* (accept a proposed offer), *Finalize* (finalize a deal with the chosen seller and vice versa), *Decline* (reject the temporarily accepted previous offer), *Withdraw* (terminate the negotiation thread). The difference between accept and finalize is that a buyer may accept several offers from multiple sellers in any one negotiation episode by making use of heuristics presents two types of agent viz. buyer and seller. We are adopting these primitives for inter-agent communication in our model.

Using game theory or heuristics approach, three strategies commonly adopted as given in [3] are: (1) conceder (2) non-conceder and (3) linear. A fourth strategy presented in implementing privacy negotiation [10] to conclude the negotiation process is also worth exploring

Additional challenges in implementing a negotiation framework include the discovery problem [8], i.e., discovery of potential agents to negotiate with. To solve this we make use of a centralized negotiation server, which will provide directory services for agent discovery.

The main purpose of every agent in Kasbah e-marketplace [4] is to finalize an acceptable deal which is in accordance with the conditions and the constraints specified by the buyers and the sellers. Unfortunately, the Kasbah agents can only negotiate over the single issue of price which is the biggest disadvantage.

Overall the following are the shortcomings of the existing negotiation protocols: (1) non-consideration of non-functional attributes (2) restricted to single issues (3) No hierarchical representation of negotiation issues. Here, we have proposed "HENRI," a new model that aims to address these issues in a computationally efficient manner.

## 3. OUR MODEL

In this paper, we propose an enhanced architecture that comprises of the negotiation system and the agents that participates in the entire process. The negotiation system implements a negotiation mechanism that uses the concept of weighted utility wherein the non-functional attributes are taken into context. The negotiation system comprises of two components: (1) advertisement repository and (2) condition-checker. The actual negotiation commences as the agents submit their advertisements into the advertisement repository. The advertisement repository is the key database that stores all the agent specific information. The control system performs conditional matching of the buyer and the sellers according of the content of the advertisements and filters them with respect to their unique product id. On satisfying the condition of the presence of at least one buyer and one seller for a particular product, the control then passes on the agents who now carry out the negotiation. The negotiation process involves the generation of offers and counter-offers for a specified slice of time and it terminates when an agreement has been established. The negotiation system is subsequently notified.

### 3.1 Negotiation System

The negotiation system has the following components: (1) advertisement repository and (2) condition checker.

The advertisement repository is constituted of the following tables: (1) Agent table: It stores the detailed information of every individual agent viz. the agent id, name, IP and type of agent (buyer or seller). (2) Product table: it contains details of products on which the negotiation is to be carried out. It contains two fields: product id and product name. (3) Attributes table: Each product has a number of attributes which can be shared in a hierarchical manner so that it can be classified into proper sub-sections and which will simplify the representation. (4) Advertisement table: It incorporates advertisements according to a specific template and formal semantics. All the necessary details that summarize are agent and the product that would be put up for negotiation. The attributes of this table are advertisement id, product id, agent id and validity counter. (5) Ongoing Negotiation table: This is a storehouse for the entire negotiation that is possessed by the agent and includes the agent id and the product id and the number of offers that have been made. The main purpose of this table is that when the negotiation process has been going on between two or more agents on a product and another agent wants to negotiate on the same product, then that agent is added directly to the same negotiation chain so as to support dynamic re-configurability.

The condition-checker inspects the advertisements in the advertisement repository and performs matching the advertisements in accordance to their product ids.

### 3.2 Agent Architecture

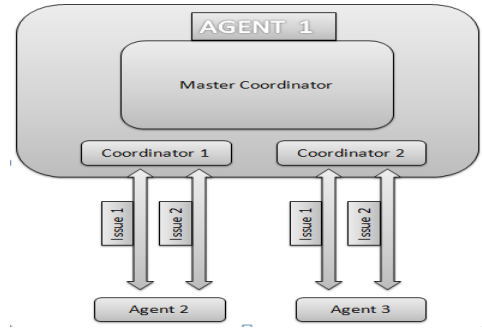

Figure 1. Agent Architecture

As shown in figure 2, the agent contains a master coordinator that monitors and controls all the coordinators present in that agent for effectively initializing and finalizing the negotiation by creating coordinator object for each agent with which it is interacting. The coordinator object creates threads for each issue and sub-issue. The coordinator is responsible for interacting with the coordinators of other agents through threads and continues the negotiation by generation of offers and counter-offers. The coordinator concurrently performs negotiation over multiple issues utilizing unique threads for each issue. If the offer generated for every single issue is agreed upon the negotiation is successful. Otherwise the threads which have been acknowledged with rejection reconsider their initial offers and regenerate their counter-offers based on their utility functions.

### 3.3 Actual Negotiation Process

The negotiation process commences with the initial offer presented by both the parties involved in the negotiation i.e., the buyer agent and the seller agent. The seller agent makes his initial offer in which it describes the maximum price of the product. The buyer agent, on the other hand, considers its offer and compares it to its initial minimum price that it has fixed for the negotiation, based on calculation of utility functions as described in the subsequent sub-section. If the prices match the two parties concur and the deal is accepted otherwise the buyer agent rejects the offer made by the seller agent and instead presents its counter-offer that lists down its minimum price.

The next round of negotiation occurs with the seller agent generating another counter-offer with reduced price. The buyer agent either accepts the deal or it again generates another counter-offer that has higher prices than its prior offer. Hence both the parties gradually converge and finally, come down to an agreement. In case of exceeding the time duration assigned for one negotiation settlement the negotiation is aborted.

During the process of negotiation, the following messages are exchanged: (1) sendAdvertisement(): intimate the negotiation system about the product. (2) sendIdentity(): provide the agent's

649

address. (3) connectThread(): create a thread for each issue on which negotiation is to be carried out. (4) offer(): send offer. (5) accept(): agree on a particular issue. (6) reject(): reject an offer on a particular issue and terminate the negotiation. (7) finalize(): finalize the negotiation.

### 3.3.1 Evaluation of utility function

For any product the non-functional attributes are those which are judgmental in determining the overall functionality of the product and do not focus on the specific issues. For example, the price (actual cost i.e., the expenses incurred for initial investments for deployment of the product and cost with margin i.e., the final price of a product generated considering the inclusion of profits over the actual cost), the ease of use, updates, time period etc. The functional attributes on the other hand, thoroughly specifies every feature and calculates their minimum and maximum utilities. (1)Minimum utilities: It is the product of all the non-functional attributes and the actual cost. (2) Maximum utilities: It is the product of all the non-functional attributes and the cost with margin. A product with multi-issues has weights i.e., priorities assigned to each issue which signifies their relative importance in the overall valuation of the product. The utility function mathematically depicts the relative usefulness of each attribute (issue) rather than specifying bounded ranges of discrete values to weights of the attributes, the user has flexibility of extending or diminishing the range of values in accordance to the requirement of particular situation. The utility is calculated as follows:

$$U_{min} = (\prod non-functional\ attributes) \times Actual\ Cost$$

$$U_{max} = (\prod non-functional\ attributes) \times Cost\ with\ margin$$

The overall utility of the product is determined by calculating the maximum and minimum payoff as follows:

$$Minimum\ Payoff = \sum U_{min}$$

$$Maximum\ Payoff = \sum U_{max}$$

For the acceptance of an offer, the offered cost of each particular issue should be greater than or equal to the minimum cost.

### 3.3.2 Generation of counter-offers

Assuming that the negotiation process commences with an offer generated by the buyer, we have an initial offer statement with included prices for each attribute. The seller inspects the offer, checking the prices of each issue in succession. It then calculates the utility of each issue with respect to the offered prices and verifies whether prices are acceptable. For unacceptable offers, the following steps are to be followed:
1. Decrease utility as follows:

$$U_{new} = U_{old}(1 - \frac{\lambda' t}{w})$$

$$where\ U_{new} = new\ utility$$

$$U_{old} = old\ utility$$
$$t = no.\ of\ rounds$$
$$w = weight$$
$$\lambda' = penalty$$

2. Derive a new value of λ(penalty according to which the utilities can be varied.
3. The coordinator is responsible for calculating the new value of λ. It takes into consideration all the attributes on which an agreement has not been reached and then depending upon number of rounds left and the weights, it derives a new value of λ as shown below:

$$\sum_{x_{i\ not\ accepted}} x_i(\lambda)$$

$$= \frac{\sum_{U_{i_{max}\ not\ accepted}} U_{i_{max}} - \sum_{U_{i_{min}\ not\ accepted}} U_{i_{min}}}{no.\ of\ rounds}$$

$$where,\ x_{i_{not\ accepted}} = cost\ of\ attribute\ i\ on\ which$$
$$agreement\ has\ not\ yet\ been\ reached$$
$$w_i = weight\ of\ attribute\ i$$

$$\lambda = proportionality\ constant$$
$$U_{i_{max}} = maximum\ utility\ of\ attribute\ i$$
$$U_{i_{min}} = minimum\ utility\ of\ attribute\ i$$

4. This value of λ is sent to the threads on which negotiation is still in progress.
5. It then seals the offers that are acceptable on a temporary settlement, while generating counter-offers for the rejected issues.
6. A time limit counter is updates after every round of negotiation. If the counter exceeds a certain limit, which has been considered as the maximum number of permissible negotiation rounds, the negotiation gets terminated.
7. Till the settlement has not been reached and the time limit counter not exceeded, the counter-offers are generated.

Before negotiation can commence, certain values have to be provided to all the agents to help them carry out the negotiation. The buyer needs values for (1) actual cost, (2) cost with margin, (3) weight of each attribute, as well as (4) some non-functional attribute which will help in calculating the utility associated with the functional attributes. On the other hand, the buyer would only know overall minimum and maximum cost. Value of the non – functional attributes may or may not be supplied to the buyer. In case it is not supplied, these values are taken from the buyer agent making a request to the negotiation system. The values of actual cost and the cost with margin are derived from the maximum and minimum cost by dividing them with number of attributes. The weights may be supplied; however, if they are not available then all the attributes are assigned the same weight. The agent might have the option of selecting different weights for attributes depending on negotiation history.

### 3.3.3 *Finalizing negotiation process*

For the entire approval of a product we need to reach an agreement on every single issue as illustrated in figure 3. For sealing the deal, both the agents i.e., the buyer and the seller, exchange finalize() messages. In case an agent is negotiating with more than one agent simultaneously, then the agreement is finalized with the agent who offers the most favorable price(minimum price for buyer and maximum price for seller).

## 4. SALIENT FEATURES

Our model presents the following salient features :( 1) Concurrent and parallel negotiation on each issue using multi-threading. (2) Multi-party and multi-issue negotiation. (3) For the generation of offers and counter-offers, the weight of each issue and utility function are taken into consideration. (4) The non-functional attributes are utilized in the evaluation of varying range of utilities. (5) Enhancements such as hierarchical alignment of the issues in restriction of the negotiation instances to a definite time limit have been implemented. (6) Addition of validity counters to advertisements to kill the advertisements on failure to start negotiation with the specified maximum time interval.

## 5. SIMULATION

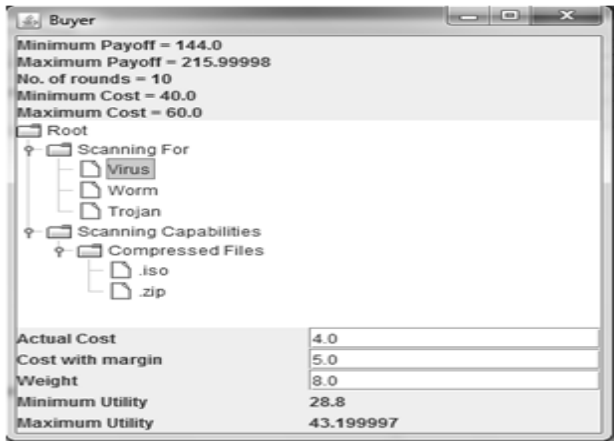

**Figure 2. Buyer Agent**

Our prototype model was developed and designed using Java NetBeans IDE 6.9.1. The Java API for XML messaging (JAXM) was used to send the messages over the network in accordance to the SOAP method calls.

In the proposed model, the negotiation container, the advertisement repository and the condition checker are all built based on the web service standards. A model of such type supports agents coming from heterogeneous environments i.e., various computer platforms and different programming languages to communicate and negotiate with each other through the application of simple SOAP protocol.

In our simulation experiment, we're considering a negotiation process involving an antivirus product. Let us assume a B2B negotiation scenario in which organization A requires an anti-virus product with certain specifications and organization B offers a product with these matching specifications. Both of them send their advertisements through their agents to the advertisement repository. Following matching of advertisements by the condition-checker, the negotiation server notifies the agents about each other's location to enable them to directly carry out negotiation between themselves.

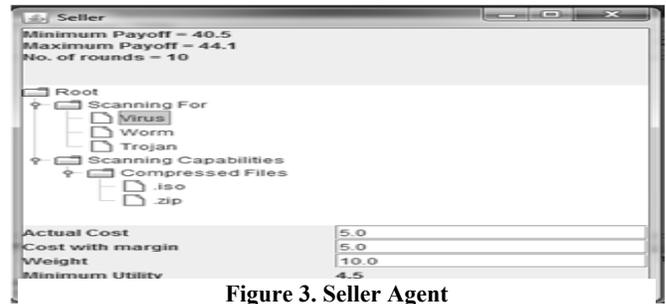

**Figure 3. Seller Agent**

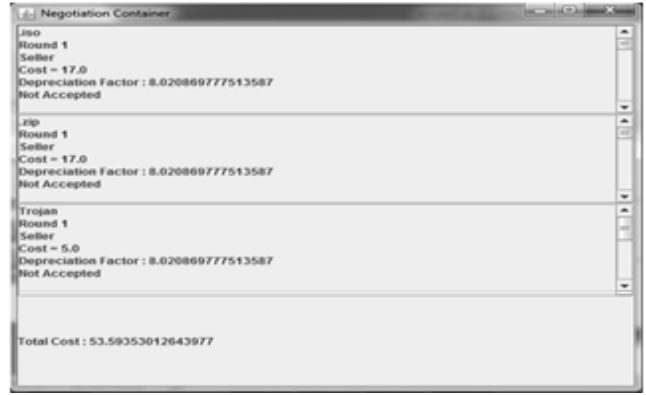

**Figure 4. Negotiation Container**

The focus now shifts to the agents who carry out the negotiation on behalf of their organizations. The seller agent (representing organization B) provides the description of each issue and its respective sub-issues in hierarchical fashion as shown in the Figure 4 together with this it also gives information about non-functional attributes for each issue viz. its actual cost, the cost with margin, its weight, updates and scope of protection.

The maximum and minimum utility is calculated by manipulating the non-functional attributes. The maximum number of rounds in this case is 10.

**The Seller side:**

The non-functional attributes are as follows:

(1) Actual Cost = 15.0 (2) Cost with margin = 17.0 (3) Weight = 8.0 (4) Updates = 1 (5) Scope of protection = 0.9

Minimum Utility = Actual Cost x Weight x Updates x Scope of protection = 15.0 x 8.0 x 1 x 0.9= 108.0.

Maximum Utility = Cost with margin x Weight x Updates x Scope of protection:= 17.0 x 8.0 x 1 x 0.9= 122.39994

Minimum Payoff= $\sum U_{min}$ as shown in section 3.3

$$= 351.0$$

Maximum Payoff= $\sum U_{max}$ as shown in section 3.3

$$= 379.8$$

**The Buyer side:**

In case of the buyer (representing organization A), it is only informed of the minimum and maximum cost and not of each issue individually. Buyer guesses the weight of the each attribute and calculates the actual cost and cost with margin of each attribute by dividing the minimum cost and the maximum cost by number of attributes. The user interface for the buyer is shown in Figure 5.

In our experiment, for the sake of simplicity, we're assigning the weight of each attribute as 8.0. We will calculate the rest of the values in the same mannerism.

Final values: (1) Actual cost= 4.0 (2) Cost with margin= 6.0 (3) Weight= 8.0 (4) Minimum Utility= 28.8 (5) Maximum Utility= 43.199997 (6) Minimum payoff= 144.0

    Maximum payoff= 215.99998

Simulating the negotiation process-We're negotiating on five issues simultaneously. Let us discuss one issue in particular. Consider Round 1: Issue=Compressed File Sub-issue= .iso

The Depreciation factor i.e., proportionality constant is calculated as shown in section 3.4.2 which equates to 8.02087.

The seller sends this initial offer to the negotiation container in the form of messages. The negotiation container acknowledges the messages and displays it at the buyer side.

In case the buyer returns a 'null' in response to this offer that alerts the negotiation container to terminate the deal. Otherwise, if the buyer messages the container with a counter-offer and this gets displayed on the seller's side again, then the next round of negotiation commences. The process continues a satisfactory deal is established or the maximum number of rounds hasn't been reached after which negotiation either succeeds of fails.

This procedure is followed for all the issues and at the acceptance of every issue, the price of them are all summed up to result into the final cost, as shown in the figure 4.

## 6. FUTURE WORK

This negotiation model has the scope of getting implemented to form into a full-fledged autonomous negotiation system. It can also provide pertinent support in identifying the behavioral patterns of the negotiating agents and classifying them into specific groups of agent's viz. conceder, linear, tough etc. The generation of counter offers requires further mathematical treatment. Currently, it is affiliated to a strict linear progression and can have the ability for switching its strategy dynamically and follow other mathematical progressions such as the geometric.

## 7. CONCLUSION

This paper presents an enhanced negotiation model which performs concurrent and parallel negotiation utilizing the concepts of multithreading thus allowing multi party multi issue negotiation. Utilization of non functional attributes in the evaluation of the relative usefulness of the attribute has been implemented. Hence this paper paves the way towards an improved approach of the negotiation scenario.

## 8. REFERENCES

[1] N. R. Jennings, P. Faratin, A. R. Lomuscio, S. Parsons, C. Sierra and M. Wooldridge, 2000.*Automated Negotiation: Prospects, Methods and Challenges. Int Journal of Group Decision and Negotiation*. GDN2000 Keynote Paper.

[2] XianrongZheng, Patrick Wendy Powley,KathrynBrohman, 2010.*Applying bargaining game theory to Web Service Negotiation*. 2010 IEEE International Conference on Services Computing.

[3] Thue Duong Nguyen, Nicholas R. Jennings, 2003.*A Heuristic Model for current bi-lateral negotiations in incomplete information settings*. Proceedings of International Joint Conferences on Artificial Intelligence, Mexico.

[4] Raymond Y.K. Lau, 2007.*Towards a web services and intelligent agents-based negotiation system for B2B eCommerce*. Electronic Commerce Research and Applications 6 (2007) 260-273.

[5] StanislavPokraev, ZlatkoZlatev, Rogier Brussee1, Pascal van Eck, 2004.*Semantic Support for Automated Negotiation with Alliances. 6th International Conference on Enterprise Information Systems*, ICEIS 2004, 14-17 April 2004, Porto, Portugal.

[6] Jin Baek Kim, ArieSegev, AjitPatankar, Min Gi Cho,2003.*Web Services and BPEL4WS for Dynamic eBusiness Negotiation Processes*. 1st International Conference on Web Services (ICWS '03), (Las Vegas).

[7] Jamal Bentahar, ZakariaMaamar, DjamalBenslimane, Philippe Thiran,2007.*An Argumentation Framework for Comunities of Web Services*. IEEE Computer Society Vol. 22, No. 6, November/December 2007.

[8] Minghua He, Nicholas R. Jennings, Ho-Fung Leung, 2003.*On Agent-Mediated Electronic Commerce. IEEE Transactions on Knowledge and Data Engineering*, vol. 15, no. 4, pp. 985-1003, July-August 2003.

[9] ZlatkoZlatev, NikolayDiakov, StanislavPokraev, 2004.*Construction of Negotiation Protocols for E-Commerce Applications*. ACM SIGecom Exchanges, Vol 5, No. 2, November 2004.

[10] SorenPreibusch, 2005. *Implementing Privacy Negotiations in E-Commerce*. CEC 2005. Seventh IEEE International Conference on E-Commerce Technology, pp. 387-390, July 2005.

[11] Chai, K. and Potdar, V., 2009. User contribution measurement model for web-based discussion forums. In: *3rd IEEE International Conference on Digital Ecosystems and Technologies*, 2009 (DEST 2009). Istanbul, Turkey, June 1-3.

[12] Chai, K., Potdar, V., and Chang, E., 2007. A Survey of Revenue Models for Current Generation Social Software's Systems. In: Proceedings of the *International Conference on Computational Science and Its Applications* (ICCSA 2007). Kuala Lumpur, Malaysia, October 24-26.

[13] Chai, K., Potdar, V., and Chang, E., 2007. A survey of revenue sharing social software's systems. In: *Proceedings of international workshop on social interaction and mundane technologies* (simtech), Melbourne,Austrlia.